\documentclass[conference]{IEEEtran}
\usepackage{times}

\usepackage[numbers]{natbib}
\usepackage{multicol}
\usepackage[colorlinks=true, citecolor=magenta]{hyperref}

\usepackage[textsize=scriptsize]{todonotes}

\newcommand{\smtr}{\textsc{SocialGym 2.0}\xspace}
\usepackage{color}
\usepackage{colortbl}
\newcolumntype{a}{>{\columncolor{green}}c}
\newcolumntype{b}{>{\columncolor{green!65}}c}
\newcolumntype{d}{>{\columncolor{green!40}}c}
\newcolumntype{e}{>{\columncolor{green!15}}c}
\usepackage[T1]{fontenc}
\usepackage{xspace}
\usepackage{multicol}
\usepackage{multirow}

\usepackage{amsthm}
\usepackage{soul}
\usepackage{threeparttable}
\usepackage{listings}
\usepackage{pifont}
\newcommand{\cmark}{\ding{51}}%
 
\usepackage{wrapfig}
\usepackage{amsmath, amsfonts, amssymb}   
\usepackage{mathtools}
\usepackage{pifont}%
\newcommand{\xmark}{\ding{55}}%
\usepackage{booktabs}
\usepackage{float}
\usepackage{bm}
\usepackage{subcaption}
\usepackage[font=small]{caption}

\definecolor{codegreen}{rgb}{0,0.6,0}
\definecolor{codegray}{rgb}{0.5,0.5,0.5}
\definecolor{codepurple}{rgb}{0.58,0,0.82}
\definecolor{backcolour}{rgb}{0.96,0.96,0.96}

\lstdefinestyle{mystyle}{
    frame=single,
    backgroundcolor=\color{backcolour},   
    commentstyle=\bfseries\color{codegreen},
    keywordstyle=\bfseries\color{magenta},
    numberstyle=\tiny\color{codegray},
    stringstyle=\bfseries\color{codepurple},
    basicstyle=\ttfamily\footnotesize,
    breakatwhitespace=false,         
    breaklines=true,                 
    captionpos=b,                    
    keepspaces=true,                 
    numbersep=5pt,                  
    showspaces=false,                
    showstringspaces=false,
    showtabs=false,                  
    tabsize=2
}

\definecolor{bluekeywords}{rgb}{0.13,0.13,1}
\definecolor{greencomments}{rgb}{0,0.5,0}
\definecolor{redstrings}{rgb}{0.9,0,0}

\lstdefinestyle{mystyle2}{language=Python,
showspaces=false,
showtabs=false,
breaklines=true,
showstringspaces=false,
breakatwhitespace=true,
escapeinside={(*@}{@*)},
commentstyle=\bfseries\color{greencomments},
keywordstyle=\color{bluekeywords}\bfseries,
stringstyle=\color{redstrings},
basicstyle=\ttfamily
}

\begin{document}

\title{\smtr: Simulator for Multi-Agent Social Robot Navigation in Shared Human Spaces}

\author{Zayne Sprague, Rohan Chandra, Jarrett Holtz, Joydeep Biswas \\{\small\{\textit{zaynesprague, rchandra, jaholtz, joydeepb\}@utexas.edu}}\\ {\small University of Texas, Austin}\\{\small Project (including code, videos, and documentation) hosted at \href{https://amrl.cs.utexas.edu/social_gym/}{\underline{\textbf{\smtr}}}}}



%

\maketitle

\begin{abstract}

We present \smtr, a multi-agent navigation simulator for social robot research. Our simulator models multiple autonomous agents, replicating real-world dynamics in complex environments, including doorways, hallways, intersections, and roundabouts. Unlike traditional simulators that concentrate on single robots with basic kinematic constraints in open spaces, \smtr employs multi-agent reinforcement learning (MARL) to develop optimal navigation policies for multiple robots with diverse, dynamic constraints in complex environments. Built on the PettingZoo MARL library and Stable Baselines3 API, \smtr offers an accessible python interface that integrates with a navigation stack through ROS messaging. \smtr can be easily installed and is packaged in a docker container, and it provides the capability to swap and evaluate different MARL algorithms, as well as customize observation and reward functions. We also provide scripts to allow users to create their own environments and have conducted benchmarks using various social navigation algorithms, reporting a broad range of social navigation metrics.

\end{abstract}
\section{Introduction}
\label{sec: introduction}


For autonomous agents to be successfully deployed in environments with human populations, it is essential to incorporate principles of collaboration and social compliance in those agents. These principles are particularly relevant in applications such as warehouse management~\cite{warehouse}, delivery of medication, provision of companionship, and navigation assistance in airports~\cite{airport}. The challenge in developing socially compliant behavior for these scenarios lies in the diversity of environments and unpredictable human traffic patterns, which require extensive training for the agents to operate safely and effectively. Deploying untrained agents in social environments is not feasible, highlighting the need for realistic simulated environments for training and testing. Such simulations should ideally mimic human navigational patterns to enhance the training and testing process for social agents.

\begin{figure}[t]
    \centering
    \includegraphics[width=\columnwidth]{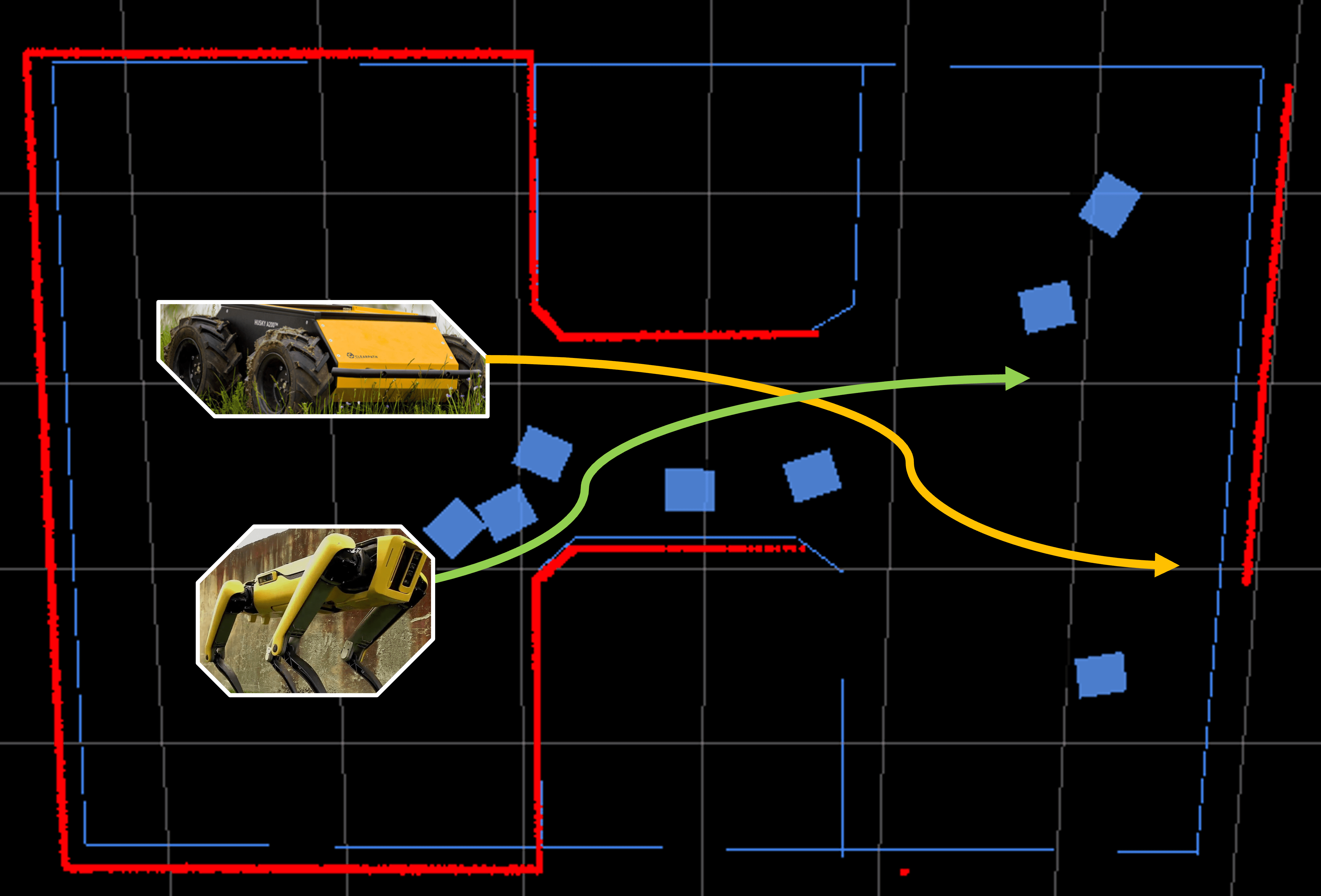}
    \caption{\textbf{\smtr} is a multi-agent navigation simulator for social robot navigation. \smtr builds on top of the PettingZoo~\cite{pettingzoo} multi-agent reinforcement learning library and interfaces with a low-level planner capable of global path planning and trajectory optimization for multiple agents with varying dynamic constraints. In this figure, blue boxes represent agents (robots or humans).}
    \label{fig: cover}
    \vspace{-10pt}
\end{figure}

Several simulators, listed in Table~\ref{tab: related_simulators}, focus on emulating specific challenges in social navigation for agents to train on. PedsimROS~\cite{social_forces} is provided as a native Robot Operating System (ROS) package that can be easily integrated into any higher-level navigation interface. SEAN 2.0~\cite{tsoi2022sean} and CrowdBot~\cite{crowdbot} use Unity~\cite{juliani2018unity} to provide a photo-realistic 3D physics engine, allowing for robot dynamics to be included in the agents physics (helping to close the sim-to-real gap during deployment). SocNavBench~\cite{biswas2022socnavbench} captures realistic human traffic patterns by replaying trajectories from popular pedestrian datasets. CrowdNav~\cite{crowdnav} focuses on dense crowd simulations for agents to navigate through.  MengeROS~\cite{aroor2017mengeros} offers several collision avoidance modules, including ORCA and social forces~\cite{social_forces} for simulating human pedestrian motion and can simulate up to $1000$ pedestrians and $20$ robots in the order of milliseconds. These simulators provide various metrics to evaluate socially compliant trajectories, including successful navigation, path smoothness, stopping time, and jerky movement.
\begin{table*}[t]
    \centering
    \resizebox{\textwidth}{!}{
    \begin{threeparttable}
    \begin{tabular}{rcccccccc}
    \toprule
         &  & & & \multicolumn{4}{c}{\textbf{CONFIGURABILITY \& EXTENSIBILITY}}\\
   \cmidrule{5-8}
          \multirow{2}{*}{Simulator}& \textbf{MULTI-AGENT}$^{\dagger\dagger}$ & \textbf{CONSTRAINED}$^\mathsection$   & \textbf{ROBOT}   &\multirow{2}{*}{Agent}&Local  & \multirow{2}{*}{Policy} & \multirow{2}{*}{Environment}\\
          & \textbf{PLANNING} & \textbf{ENVIRONMENT} &\textbf{DYNAMICS}  & & Navigation & &\\
     \midrule
\multirow{2}{*}{SEAN~\cite{tsoi2022sean}}&\multirow{2}{*}{\xmark}&\multirow{2}{*}{\xmark} &\multirow{2}{*}{\cmark}&\multirow{2}{*}{\xmark}&Obstacle avoidance,&\multirow{2}{*}{Training, Evaluation}&\multirow{2}{*}{Scenarios}\\
&&&&&Trajectory planning&&\\

\rowcolor{lightgray}CrowdBot~\cite{crowdbot}&\xmark&\xmark&\cmark&Kinodynamics, Sensors&\xmark&\xmark&\xmark\\

SocNavBench~\cite{biswas2022socnavbench}&\xmark&\xmark&\cmark$^\ddagger$ &Sensors&\xmark&\xmark &2d maps, Scenarios\\    

\rowcolor{lightgray}CrowdNav~\cite{crowdnav}&\xmark&\xmark&\xmark&\xmark&\xmark&Training, Evaluation&\xmark \\

\multirow{2}{*}{MengeROS~\cite{aroor2017mengeros}}&\multirow{2}{*}{\xmark}&\multirow{2}{*}{\xmark}&\multirow{2}{*}{\xmark}&\multirow{2}{*}{Type, Sensors}&Obstacle avoidance,&\multirow{2}{*}{\xmark}&\multirow{2}{*}{Crowd configurations}\\
&&&&& Trajectory planning& &\\

\rowcolor{lightgray}PedsimROS~\cite{social_forces}&\xmark&\xmark&\xmark&\xmark&\xmark&\xmark&Crowd configurations\\

\multirow{2}{*}{SocialGym~\cite{socialgym}}&\multirow{2}{*}{\xmark}&\multirow{2}{*}{\xmark}&\multirow{2}{*}{\cmark}&Type, Sensors&Obstacle avoidance,&\multirow{2}{*}{\xmark}&\multirow{2}{*}{\xmark}\\
&&&&Kinodynamics& Trajectory planning&&\\

\midrule

\multirow{2}{*}{\textbf{\smtr}}&\multirow{2}{*}{\cmark}&\multirow{2}{*}{\cmark}&\multirow{2}{*}{\cmark$^\star$}&Type, Behavior,& Obstacle avoidance,&Training, Evaluation, &2d maps, Scenarios,\\
&&&&Kinodynamics, Sensors& Trajectory planning&Evaluation metrics &Navigation graphs\\

     \bottomrule
    \end{tabular}
    {\footnotesize\begin{tablenotes}
\item[$\dagger\dagger$]Each agent follows a policy designed to maximize \textit{their individual} reward. This excludes crowd simulation models such as ORCA~\cite{orca} and Social forces~\cite{social_forces}.
\item[$\mathsection$] Constrained environments refer to social mini-games \textit{e.g.} Doorway, Hallway.
\item[$\star$] Different robots can simulate varying configurable dynamics.
\item[$\ddagger$] Limited to unicycle kinematic model.
\end{tablenotes}}
\end{threeparttable}
    }
    \caption{Comparison of current simulators for social robot navigation.}
    \label{tab: related_simulators}
    \vspace{-10pt}
\end{table*}

However, simulators can often lack in specific ways that limit their ability to model challenging unstructured social scenarios in the real world. For example, multiple agents act autonomously, achieving their own objectives rather than following a fixed crowd simulation model such as Pedsim or Social Forces~\cite{social_forces}. Furthermore, often such multi-agent interactions are non-cooperative or competitive, resulting in deadlocks or near-collisions~\cite{fridovich2020efficient,le2022algames}. Lastly, agents in the real world obey complex kinodynamic constraints.

In Table~\ref{tab: related_simulators}, we summarize the state-of-the-art social navigation simulators. An immediate observation is that \textit{all} simulators are currently single-agent navigation in simple open environments. To simulate crowds or other agents, the simulators will often model human crowds using reciprocal policies~\cite{orca} or replay stored trajectories from a dataset~\cite{biswas2022socnavbench}, or both. SEAN 2.0~\cite{tsoi2022sean} defines social navigation scenarios via social maneuvers such as crossing, following, and overtaking, but these only apply in open environments, excluding geometrically constrained scenarios. Crossing or passing may be impossible and lead to sub-optimal trajectories like colliding with walls when navigating through a narrow doorway, for example. Furthermore, while several simulators~\cite{tsoi2022sean,biswas2022socnavbench,crowdbot,socialgym} model real-world robot dynamics and kinematics realistically, only two simulators (CrowdBOT and our previous work, SocialGym) allow configurability and extensibility to experiment and benchmark different robot kinodynamic configurations. In fact, we find that this configurability and extensibility are desirable properties of all features in a simulator. Current simulators, however, offer users very little control over the simulator through the use of the convention-over-configurability~\cite{chen2006convention} design philosophy.
To overcome these challenges we introduce \smtr, an open-source simulator for multi-agent social autonomous navigation in challenging environments. \smtr features multiple autonomous agents, each optimizing its own objective function. Each agent is a robot with realistic kinodynamic constraints, including limits on linear and angular velocity and acceleration and physical parameters like shape and size. \smtr simulates both open, building floorplans, as well as social encounters like hallway passing. Finally, we offer users complete control over each part of the simulator, enabling users to conduct research in agent modeling, trajectory planning and collision avoidance, policy learning, and navigation in different social contexts. The unique novelty of \smtr is that it \textit{simultaneously} goes-- 

\begin{enumerate}

    \item \textbf{beyond motion replay and reciprocity:} Our multi-agent reinforcement learning paradigm trains multiple autonomous agents each with their own policy. Using the PettingZoo~\cite{pettingzoo} (multi-agent Gym) and Stable Baselines3~\cite{sb3} APIs, \smtr supports multi-agent reinforcement learning, configurable observation and reward functions, and variable number of agents across training episodes.
    \item \textbf{beyond simple kinematics:} \smtr implements global path planning and local trajectory optimization conditioned on real robot dynamics. In \smtr, robot dynamics can be configured to simulate multiple different robots with varying dynamics including differential drive and omni-directional robots dynamics. 
    \item \textbf{beyond open spaces:} \smtr simulates challenging environments including university campus buildings and geometrically constrained social mini-game scenarios. 
    \item \textbf{beyond convention-over-configuration paradigms:} \smtr uses the \textit{configuration}-over-\textit{convention} paradigm providing users control over every module of the stack, while \textit{simultaneously} keeping the stack simple to use. 
    
\end{enumerate}

\section{Background}
\label{sec: related_work}

Having robots navigate in shared humans spaces is a central goal in robotics. The core challenges in solving this problem stem from a single fact--robots have to interact with humans in shared constrained environments, which often means attempting to optimize individual conflicting objectives, such as trying to pass through a narrow hallway or doorway together. In particular, the first challenge to a robot would be to account for the hidden objectives of humans; more specifically, the optimal actions of a robot would be dependent on the unknown goals of the humans in the scenario. MARL has shown great promise to address the first challenge in many fields of robotics and engineering. The next challenge is to plan trajectories that are not only safe and efficient but more crucially are also socially compliant. Ensuring all these qualities in the resulting motion planning requires modeling precisely the underlying kinodynamics of the robots. Finally, humans are different and move in different ways according to culture, situation, and behavioral disposition. Modeling human-robot interaction is necessary to capture this range of behaviors. In what follows, we expand on recent work along MARL, local trajectory planning and robot dynamics, and human-robot interaction that motivated the design choices in \smtr.

\subsection{Multi-agent Reinforcement Learning}

MARL is a field of study with a focus on computing an optimal policy for multiple agents using reinforcement learning. Deep learning-based MARL\footnote{henceforth simply referred to as MARL.} has achieved remarkable success in cooperative, competitive, and mixed games such as Go~\cite{alphago}, chess~\cite{chess}, poker~\cite{poker}, Dota2~\cite{berner2019dota}, and StarCraft~\cite{vinyals2017starcraft}, the latter two also serving as benchmarks for fostering MARL research. In all the above, the best MARL policies even beat professional players in all these games. Recently, MARL was also applied to competitive racing and was able to defeat professional human racers~\cite{gtsophy}. We refer interested readers to \cite{gt-marl-survey} for a survey on MARL in games.

MARL has also been successfully applied to robot navigation in both indoor and outdoor scenarios. In the outdoor scenarios, several works use neural networks to either directly learn a navigation policy~\cite{weerakoon2022terp, sathyamoorthy2022terrapn, weerakoon2022graspe, guan2022ga} or learn the underlying dynamics~\cite{karnan2022vi} . For indoor navigation, Frozone~\cite{sathyamoorthy2020frozone} and DenseCAvoid~\cite{sathyamoorthy2020densecavoid} addresses the freezing robot problem in dense crowds, CoMet~\cite{sathyamoorthy2021comet} attempts to learn group cohesion to navigate among groups of pedestrians, and CADRL~\cite{cadrl}, or Collision Avoidance with Deep Reinforcement Learning, is a state-of-the-art motion planning algorithm for social robot navigation using a sparse reward signal on reaching the goal and penalises agents for venturing close to other agents. A variant of CADRL uses LSTMs to select actions based on observations of a varying number of nearby agents~\cite{cadrl-lstm}. CADRL, however, uses a unicycle kinematic model and does not account for robot dynamics. The progress being made on social navigation through MARL algorithms inspired us to include it in \smtr. Furthermore, \smtr currently supports indoor navigation but can be extended to outdoor scenarios via global vector maps.

\subsection{Robot Dynamics and Trajectory Optimization}

The goal of autonomous navigation is to move from place A to place B with little to no human input. However, solving the navigation problem requires that the resulting trajectories be not only feasible but also smooth and admissible to the low-level motion controller. Planning algorithms that ignore robot dynamics or assume simplified kinematics require non-trivial and often expensive post
processing to make the path smooth and admissible to the
controller~\cite{park2012robot}.

Simplified dynamic models assume that robots only operate in a limited subspace of their entire state space, such as low acceleration and speed, minimum wheel slip, negligible tire deformation, and perfect non-holonomic constraints~\cite{atreya2022high}. In social situations, humans execute a wide range of dynamic behaviors like slowing down to let others pass or speeding up to overtake a group of pedestrians on a sidewalk. Realistically simulating and deploying social maneuvers on robots in shared humans spaces requires that the motion planning satisfy the dynamic constraints of the robot such as linear velocity, acceleration, and steering angle~\cite{atreya2022high,karnan2022vi,wei2022steady}.  In addition to feasibility and smoothness, resulting motion controllers also need to generate optimal, efficient, and socially compliant paths. In \smtr, users can configure every kinodynamic variable and directly test the changes on the local planner. 

\subsection{Human-Robot Interaction}

Humans navigate differently in varying social contexts, such as doorways, hallways, intersections, roundabouts, and so on. The differences being the way people interact in shared spaces which engenders a range of maneuvers like passing, overtaking, following, and cutting off individuals~\cite{chandra2022gameplan, chandra2020cmetric, chandra_thesis, gameopt, socialmapf}. By simulating different scenarios such as the ones mentioned above, researchers can understand the complexities of human behavior and design robots that can navigate these scenarios effectively and safely~\cite{socialnav_survey,stone_nav_survey}. \smtr supports open spaces and building floorplans, enabling the modeling of macroscopic crowd patterns as well as microscopic social interactions like social mini-games (doorways, hallways, intersections, and roundabouts).

\section{The \smtr Design \& Architecture}
\label{sec: architecture}

\begin{figure}[t]
    \centering
\includegraphics[width=\columnwidth]{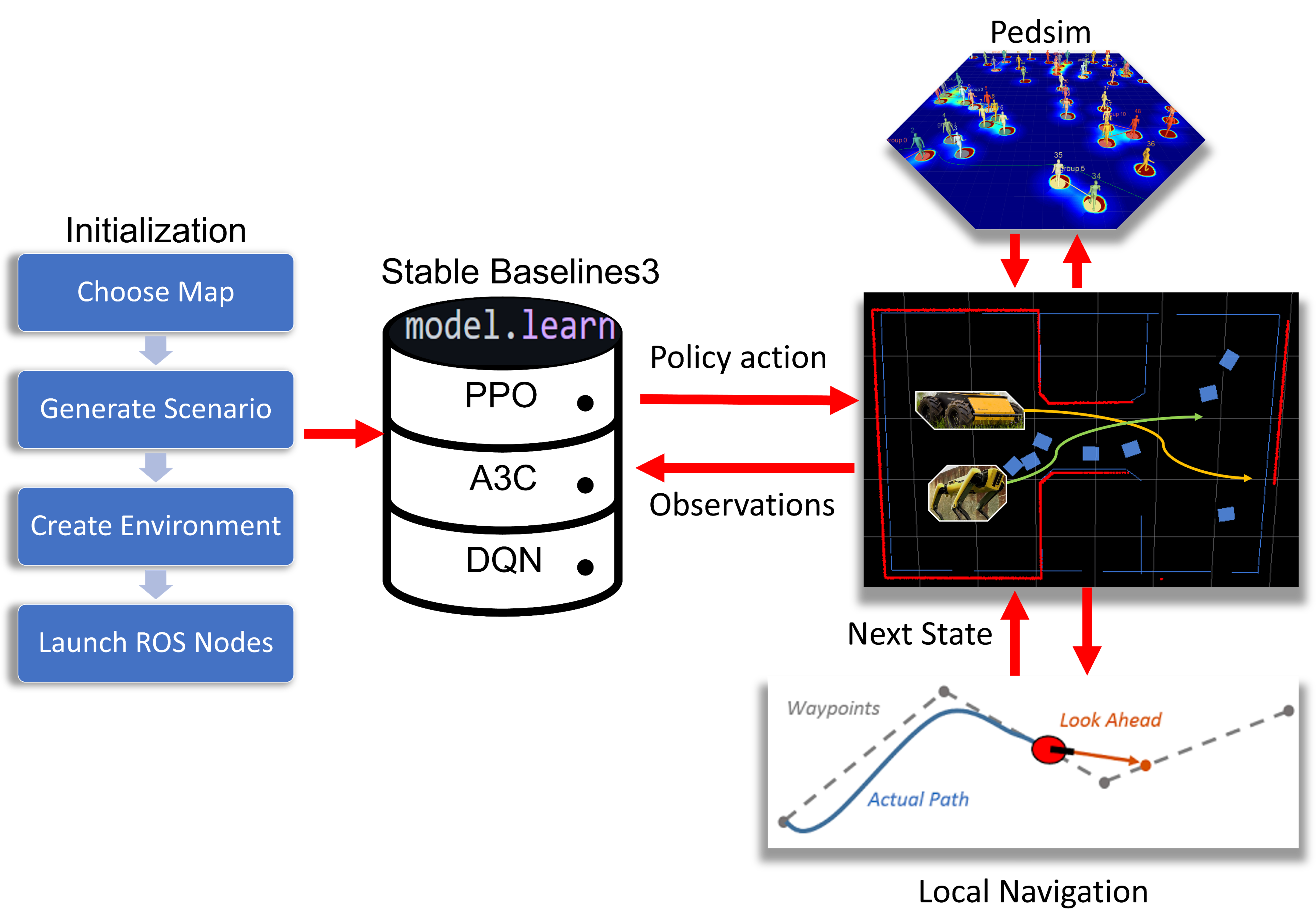}
    \caption{\textbf{\smtr Architecture Overview:} The top-level interface consists of a Pettingzoo~\cite{pettingzoo} \texttt{environment} and acts as the primary interface between the learning agents (policy) and simulator. This interface updates the agents' action selection policy based on the current observation in the state space and sends new actions to the local simulator, which returns a new state in the state space by coordinating with the Human and Navigation modules to simulate the transition function based on the selected action and robot dynamics at each time step. The new state observations are returned to the top-level interface for computing the rewards and updating the policy.}
    \label{fig: overview}
    \vspace{-10pt}
\end{figure}

In this section, we overview the different components of \smtr. We will begin by summarizing the overall design and how the different components interface with one another. In the remainder of the section, we will dive into each component in detail. 

We developed \smtr keeping configurability, extensibility, and modularity in mind, using a configuration-over-convention style. In order to allow easy development and research on various aspects of social navigation, we stratified \smtr's stack into different layers shown in Figure~\ref{fig: overview}. At the top of the stack is the PettingZoo~\cite{pettingzoo} and Stable Baselines3~\cite{sb3} interface. This interface uses ROS to send actions from a policy to UTMRS\footnote{University of Texas Multi-Robot Simulator}, a lightweight simulation engine that acts as an intermediate between the interface and the local navigation and the human crowd simulation modules. The local navigation planner is responsible converting high-level actions from the PettingZoo interface into continuous motion commands that satisfy the underlying robot dynamics and sends back the next state to the simulation engine. Each layer of the stack has a modular API that allows researchers and developers to focus on a single part of the stack at a time without having to refactor or access other parts of the stack. In the sections that follow, we describe each part of the stack in detail.

\subsection{The Multi-Agent Gym Interface}
The top level interface follows from the familiar OpenAI Gym API extended for multi-agent scenarios using the PettingZoo and Stable Baselines3. We construct an environment that follows the standard lifecycle of a Gym environment (reset, step, etc.). Our environment takes as input a 2D map which consists of a vector map file containing vectors that represent walls (or otherwise impassable and stationary objects) as well as a navigation graph which defines the possible paths through the vector map.  We provide a useful program for creating 2D maps~\cite{Vector_Display}. Once a 2D map is selected, the user can choose a scenario that consists of unique starting and ending positions of agents or simulated pedestrians. To better understand the difference between a 2D map and a scenario, consider a doorway as a 2D maps with two scenarios--one in which all agents are entering and exiting the doorway in the same direction, and the other has agents entering and exiting from both sides. Once a 2D map and corresponding scenario have been selected, they are passed into configuration files and given to UTMRS for tracking state as well as initializing other submodules with the same information.  Although we mention only one 2D Map and scenario here, we have wrappers that allow an environment to sample multiple 2D maps and scenarios during training and evaluation.

\subsection{UTMRS}

UTMRS is a C++ simulation \textit{engine} that receives high-level actions from the multi-agent Gym interface and is responsible for updating the state, receiving new state observations from the local navigation and human motion submodules, and sending them back to the interface via ROS messages. UTMRS additionally creates visualizations and maintains several internal states necessary for the simulator, including walls, current positions, goal states, and previous actions. Beyond serving as an engine that controls message passing and centrally interfacing all the different components, UTMRS itself does not actively impact policy learning. Although ROS is essential to our system, we designed \smtr such that users who may be unfamiliar with ROS do not have to work with ROS in their development process--\smtr handles ROS messaging internally.

\subsection{Local navigation}
The local navigation planner serves two purposes in \smtr.  The first is to facilitate continuous navigation on the navigation graph selected in the multi-agent Gym interface.  This navigation is achieved by sampling a set of trajectories at every given state and then selecting a trajectory that is in the direction of the intermediate goal (a node on the navigation graph) which is not blocked by some obstacle (robot or wall). The second purpose of the local planner is to emulate robot dynamics for continuous state changes on the selected trajectory.  This emulation ensures the continuous navigation is admissible to any specific local motion controller.  Once the trajectory has been selected and the continuous action sampled, the local planner updates the state of the agent and returns the newly updated location of the agent to UTMRS, at which point the UTMRS layer will update its internal state and pass the state as a message to other submodules as well as the Gym interface. UTMRS then awaits a new command from the multi-agent Gym interface.  This loop of high-level actions from the Gym interface being passed into UTMRS and then to the local planner for state updates, which is then passed back to UTMRS is the main simulation loop for \smtr. The human motion module follows a similar process as the local navigation module. In \smtr, the human motion is simulated using social forces~\cite{social_forces}, similar to the current simulators listed in Table~\ref{tab: related_simulators}.  Although we support the human motion model, it is optional.  \smtr supports both single and multi-agent training as well as training with simulated humans or not.

\section{Training and Evaluating a Multi-Agent Navigation Policy in \smtr}
\label{sec: training}

Having described the individual components of \smtr in the previous section, we now walk through the process of training and evaluating a multi-agent navigation policy. The life-cycle of training a multi-agent policy in the interface follows almost exactly from the standard process in OpenAI Gym or Stable-Baselines.  We extend these loops with customizations for the UTMRS layer as well as with PettingZoo to enable multi-agent training (MARL). We show an example of the required code in Listing \ref{lst:training_loop}. Although using \smtr as you would Stable Baselines v3 or PettingZoo is supported, we also implemented a program that can run the training from a configuration file. See an example of a configuration file in Listing \ref{lst:example_config}. To illustrate \smtr's features, we briefly discuss each line of code in Listing \ref{lst:training_loop}. 

The training life-cycle begins with selecting 2D maps and scenarios (Section~\ref{subsec: 2dmaps}) to be played out on those maps. Then, two class definitions, the Observer and Rewarder, are used for tracking the observations and rewards during each step (Section~\ref{subsec: obs_and_rew}). The environment is then defined, which instantiates the Gym Environment and initializes the ROS submodules with all the information needed to load the first 2D map and scenario (Section~\ref{subsec: env}). 
Next, a Stable Baselines-v3 Policy is chosen (both SB3 and SB3-Contrib are currently supported), and the learning method can be called to train the policy. Finally, the policy is evaluated on metrics designed to measure socially compliant navigation (Section~\ref{subsec: training_eval}).  We dive into each of these steps in more detail below.

\begin{lstlisting}[language=Python, caption=Example usage of SocialGym 2.0., label=lst:training_loop]
# Creating a scenario given the 2D map folder
scenario = GraphNavscenario('envs/scenario/hallway')

# Creating the Observer through modular Observations that are customizable
observations = [
  AgentsPose(ignore_theta=True), 
  OtherAgentObservables(ignore_theta=True),
  CollisionObservation(),
  SuccessObservation()
]
observer = Observer(observations)

# Creating the Rewarder with a sparse goal reward and a penalty term that scales over the course of training.
rewards = [
  Success(weight=100),
  LinearWeightScheduler(Collisions(), duration=10_000)
]
rewarder = Rewarder(rewards)

# Create the base class
env = RosSocialEnv(observer, rewarder, scenario, num_agents=7)

#... Wrappers as needed ...

# Standard Gym Interfacing for Training and Stepping 
model = PPO("MlpPolicy", env)
model.learn(total_timesteps=10_000)

obs = env.reset()
while env.agents:
    action, _states = model.predict(obs)
    obs, rewards, terminations, infos = env.step(actions)
\end{lstlisting}

\subsection{2D Maps, Navigation Graphs, and scenarios}
\label{subsec: 2dmaps}
We posit that a big part of being socially compliant is derived from experience in various geometrically constrained and dense environments where spatial and temporal reasoning are required to avoid collisions while respecting others.  To enable this in our simulator, we created a program (deployed in Docker to ease its installation and use) to create 2D maps with two components.  Each 2D map contains a list of vectors that represent impassible obstacles (walls, for example) -- denoted in blue in Figure~\ref{fig: social_minigames}. These vectors allow us to create various ``social mini-games'' enabling rapid training and evaluation of agents under challenging situations.  However, we can also use 2D floor plans of buildings to test agents in larger and more realistic situations.  

The second component of the 2D maps is the navigation graph.  The navigation graph provides all possible paths through the environment where agents must navigate through a set of nodes by following edges that connect them.  The navigation graph allows for agents to have a high-level discrete action space, i.e., GO or STOP actions, which we currently used for our evaluations (although continuous actions are soon to be supported, which could ignore the navigation graph entirely).  Having a navigation graph also enables the easy creation of constrained paths where the agent must traverse an edge shared with another agent ensuring a conflict will happen if the agents ignore each other.  

Finally, a scenario is defined as a selected list of global paths (list of nodes on the navigation graph) for each agent in the episode.  A 2D map and navigation graph could have many scenarios defined on them (for example, unidirectional traffic or bidirectional traffic on the navigation graph).  We allow for the easy creation of scenarios through a python helper class that allows the user to define global paths for the agents in each episode. 

\subsection{Observations and Rewards}
\label{subsec: obs_and_rew}

A complexity of social navigation is the vast number of definitions and ways of representing social navigation; for example, there is no ubiquitous metric for social navigation~\cite{socialnav_survey,stone_nav_survey}. We attempt to address the ambiguities in the setup, evaluation, and definition of social navigation by making the state space and reward functions completely customizable, in addition to making the simulator open-source.  This customization allows researchers to create their own definitions of social navigation (what is observable, what is hidden, what is punished, and what is rewarded) in as few line changes as possible.  We also give an intuitive class structure that allows researchers and developers to create their own observations and reward functions, extending the features of \smtr.  To ease the complexity of these customizations and to parse the state representations returned from UTMRS, we create helper classes called Observer and Rewarder.

The Observer parses the raw state vectors given from the UTMRS layer and produces an observation vector (numpy array) as well as an observation dictionary (a map between the name of a given observation with its value).  The Observer class can be thought of as a lightweight wrapper around the numpy array traditionally given to a Stable Baseline Policy; however, the construction is entirely customizable.  Allowing researchers and developers to construct different state spaces for different mathematical models of social navigation with ease.  We give an example of such a definition for our evaluations in Section~\ref{sec: experiments}.  This construction also allows for building custom layers for representation learning. In our evaluation, we build a custom LSTM layer to collapse the variable number of agent or human observations into a fixed-sized observation vector (a method used in state-of-the-art social navigation models) \cite{cadrl-lstm}.
\begin{figure*}
    \centering
\begin{subfigure}[h]{0.19\textwidth}
    \includegraphics[width=\textwidth]{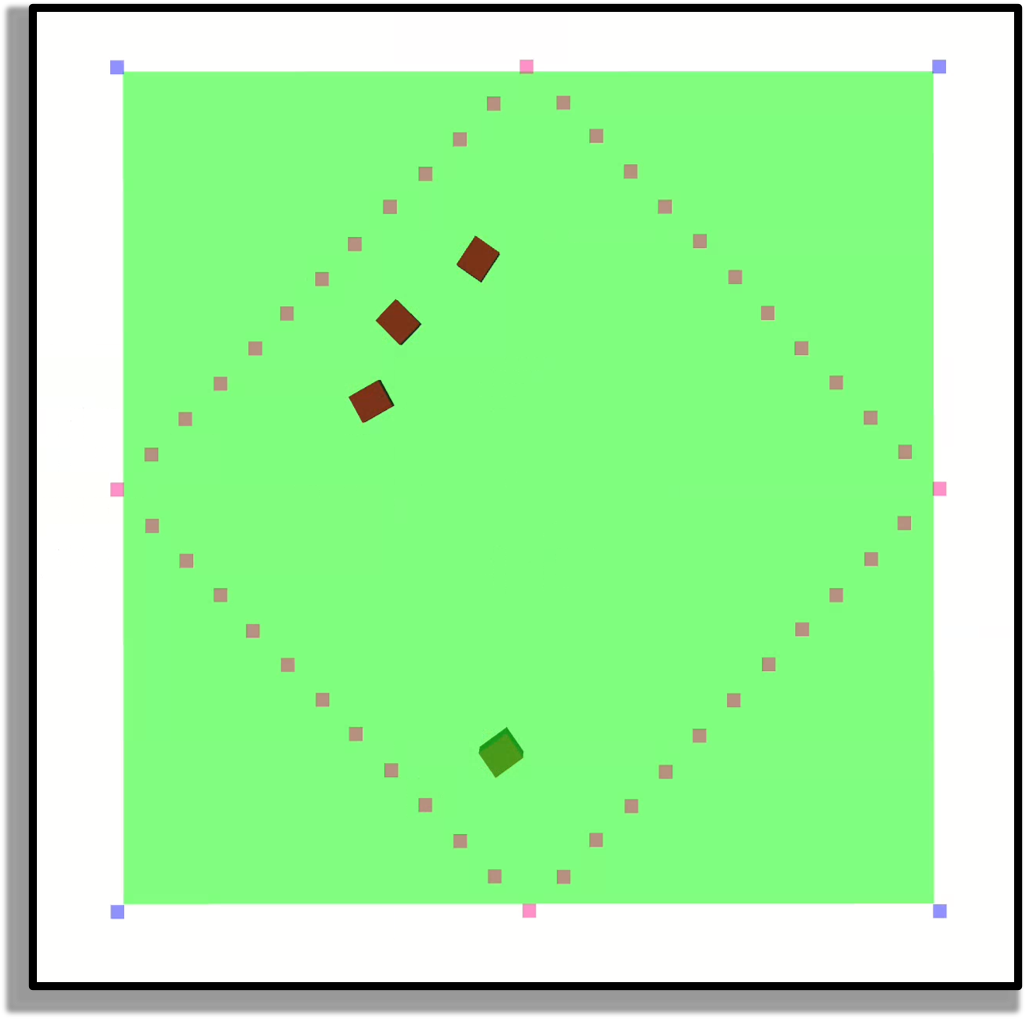}
    \caption{Open Scenario}
    \label{fig: open}
  \end{subfigure}
  \begin{subfigure}[h]{0.19\textwidth}
    \includegraphics[width=\textwidth]{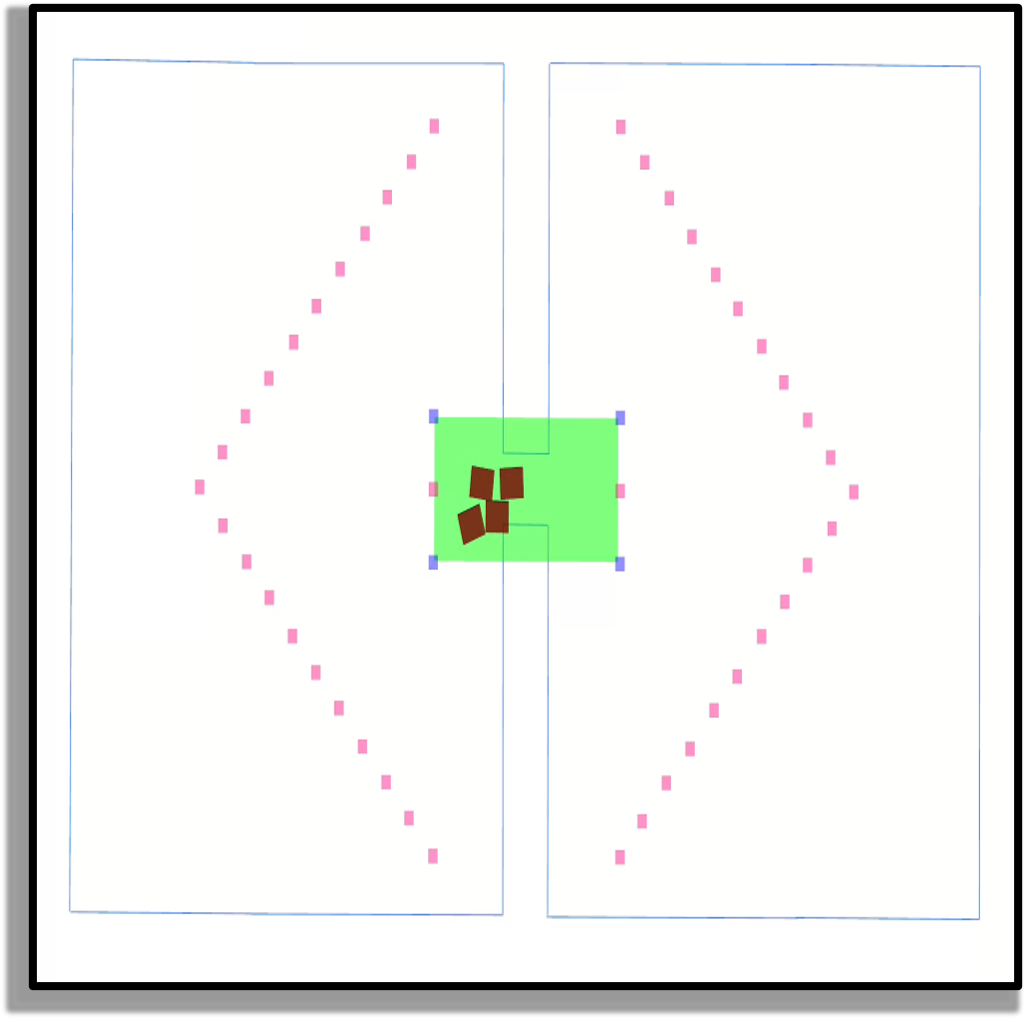}
    \caption{Doorway Scenario}
    \label{fig: doorway}
  \end{subfigure}
  \begin{subfigure}[h]{0.19\textwidth}
    \includegraphics[width=\textwidth]{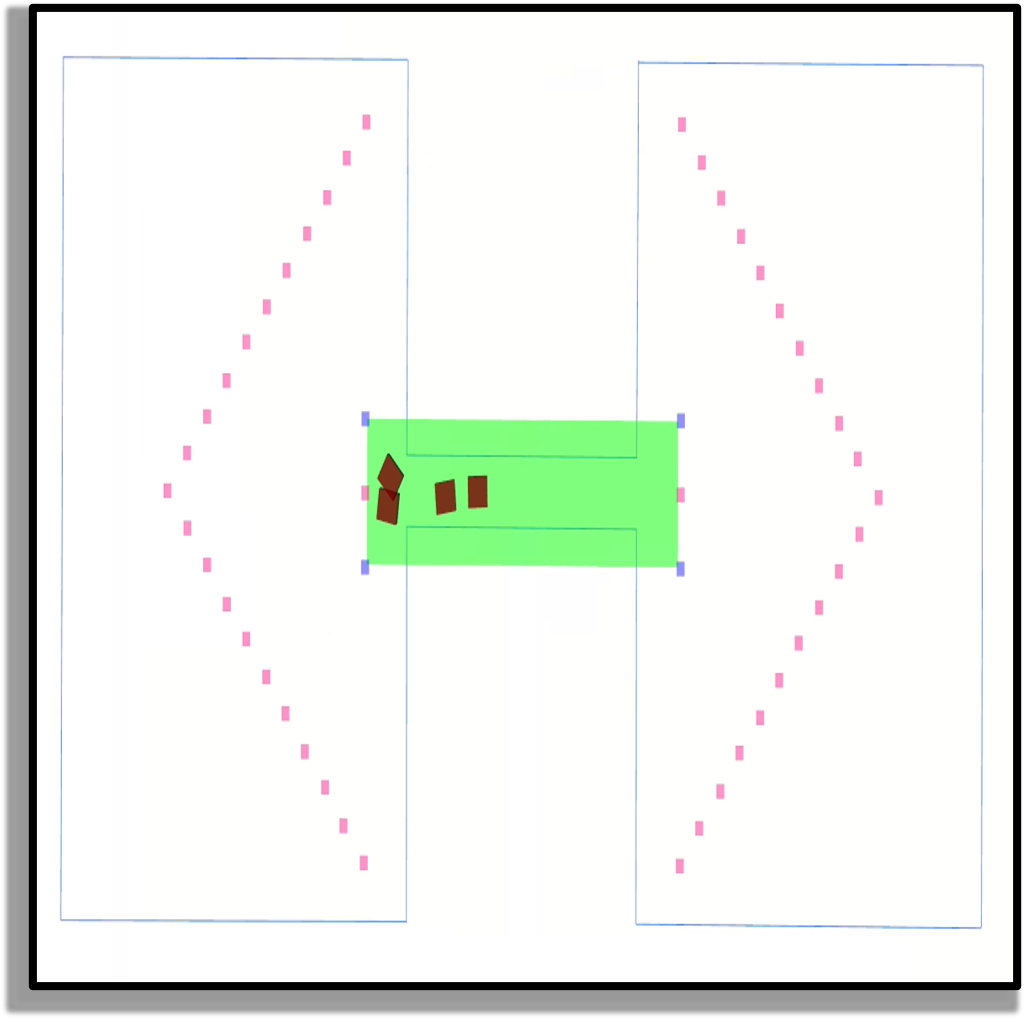}
    \caption{Hallway Scenario}
    \label{fig: hallway}
  \end{subfigure}
  \label{fig: social_minigames}
  \begin{subfigure}[h]{0.19\textwidth}
    \includegraphics[width=\textwidth]{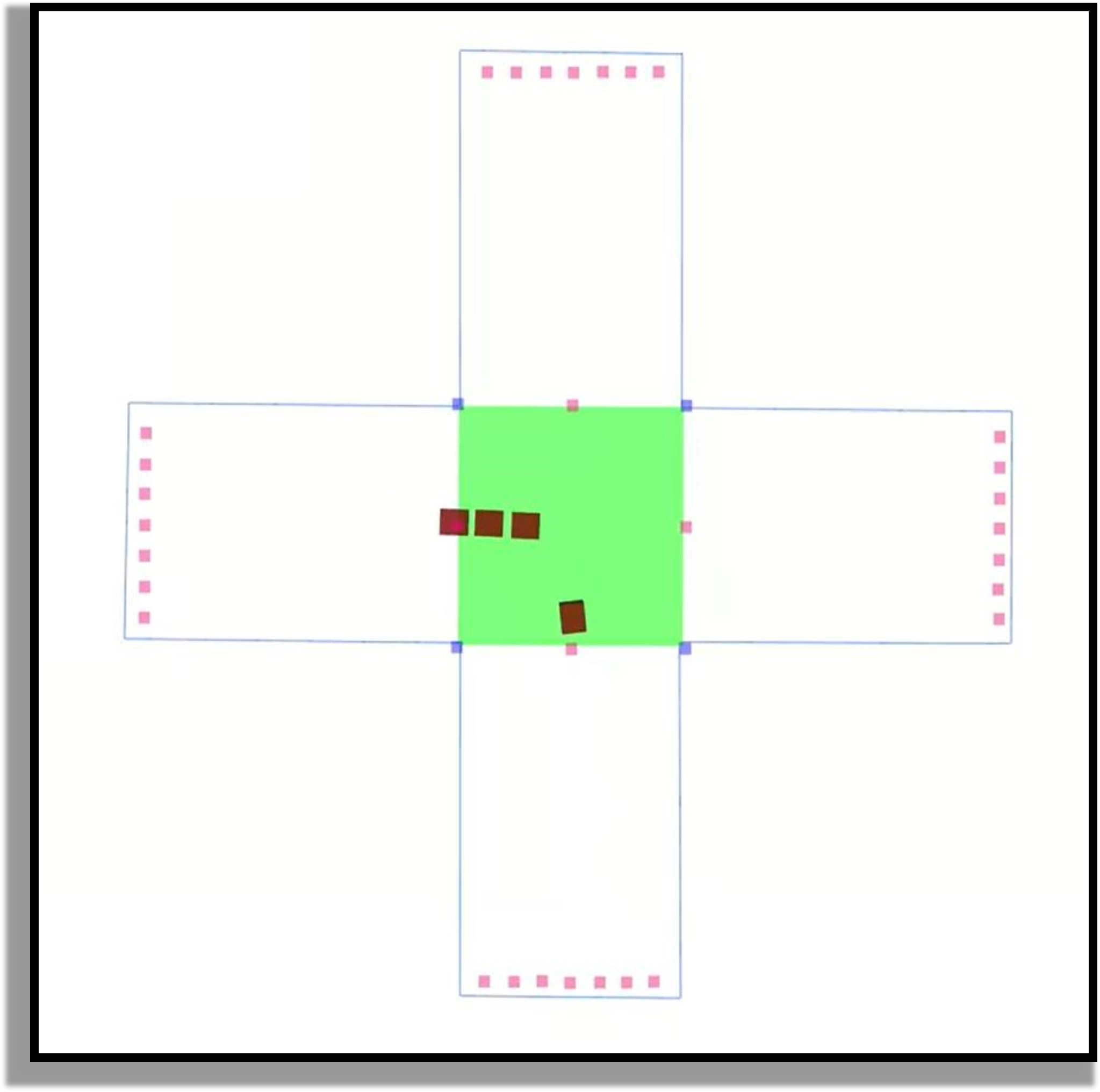}
    \caption{Intersection Scenario}
    \label{fig: intersection}
  \end{subfigure}
  \label{fig: social_minigames}
  \begin{subfigure}[h]{0.19\textwidth}
    \includegraphics[width=\textwidth]{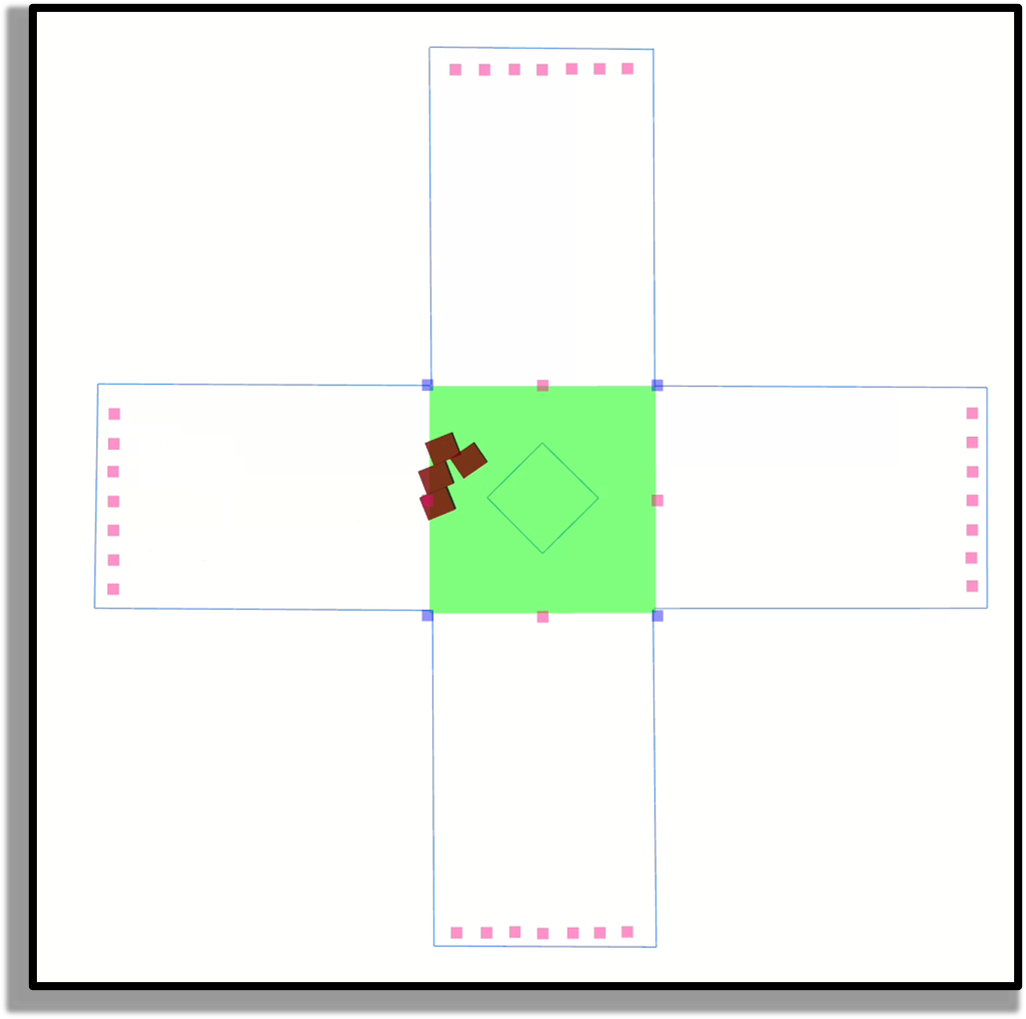}
    \caption{Roundabout Scenario}
    \label{fig: roundabout}
  \end{subfigure}
  \caption{\textbf{Social Navigation Environments:} \smtr provides the flexibility for users to create new maps and scenarios for social navigation. By default, we include five types of social environments--Open, Doorway, Hallway, Intersection, and Roundabout, in addition to the UT Austin campus buildings from SocialGym 1.0.}
  \label{fig: social_minigames}
  \vspace{-10pt}
\end{figure*}

The Rewarder class functions similarly to the Observer.  The construction takes a list of Reward base classes where each Reward class takes observations (both the vector and dictionary) from the Observer class at each step. The Rewarder can then use the observations to derive a reward or penalty for that step.  All Reward classes are summed in the Rewarder class at the end of each step; however, a dictionary of rewards is kept for logging purposes.



\subsection{Wrapping the Environment}
\label{subsec: env}
Following the Gym API, \smtr supports environment wrappers for extensibility and customization of environments as well as callbacks that tap into the life cycles of a training procedure.  \smtr has many wrappers already created to ease training including some custom MARL wrappers that end the episode when agents do not move, when agents collide, or when a step limit is reached.  Other wrappers that plot and monitor training and evaluation metrics as well as generating new scenarios that vary the number of agents, or 2D maps are also available in \smtr. We also created checkpoint and evaluation callbacks that allow the policy to be saved and tested during training.  

\subsection{Evaluation Metrics}
\label{subsec: training_eval}

In addition to the standard evaluation metrics available in Stable Baselines3 and PettingZoo, we extend these functions and implement custom evaluation metrics in the same style of SocNavBench and SEAN2.0. This stems from the noted ambigious definition of social navigation. Often, in lieu of a single metric that best defines Social Compliance, multiple metrics are used as a proxy.  We implemented the most common metrics used to measure social compliance, including partial and full success rates, velocity changes, average stopping time, collision rates, and more.

\subsection{Social Mini-game Benchmarks}
\label{sec: minigames}

In this work, we include five mini-game scenarios to benchmark social navigation. These are Open, Hallway,  Doorway, Roundabout, and Intersection, depicted in Figure~\ref{fig: social_minigames}. Social mini-games may be described as a scenario with multiple agents accomplishing a shared goal in spatially constrained environments. Such scenarios frequently arise indoors at schools, hospitals, airports, etc. as well as outdoors on sidewalks and traffic intersections. We provide a point-click/drag interface that allows for the easy construction of both vector maps and navigation graphs. A navigation graph is a collection of nodes connected by straight-line edges. An agent is then given a set of nodes to reach, where the first in the list is the starting position, and each node after should lead the agent to the last, defining a trajectory. The new custom environment can then be called easily within the top-level script, as shown on line 3 of Listing \ref{lst:training_loop}.

\section{Experiments and Discussion}
\label{sec: experiments}
In \smtr, users can represent social navigation through different formulations depending upon the application. In our evaluation, we decided to formulate social navigation as a partially observable stochastic game (POSG)~\cite{posg} with $k$ agents using the tuple, $\big \langle k, \mathcal{X},\{\mathcal{U}^i\},\mathcal{T},\{\mathcal{O}^i\},\{\Omega^i\}, \mathcal{R}^i \big\rangle$. Each agent is randomly initialized with a start position ($p^i_I$) and a goal position ($p^i_G$). The continuous state space $\mathcal{X}_t$ is an array comprising the agents' state vectors $x^i_t$. This vector is typically in SE(2) indicating that the robot has a 2D translation and an orientation. In \smtr, users can configure the state vector by adding, removing, or toggling variables. Out of the box by default, \smtr currently sets $x^i_t =  [d^i_G, p^i_x, p^i_y, \dot p^i_x, \dot p^i_y, \psi^i, v^i]^\top$ for $i = 1,2,\ldots,k$, where $d^i_G$ is the distance from the goal, $p, \dot p$ represent position and velocity, $\psi^i$ represents the heading, and $v^i$ represents the preferred velocity. 

The state space is partially observable in our formulation because agents have different goals they are trying to reach that are only known to them. This definition can be easily changed with a single line, in our case uncommenting line 3 in Listing \ref{lst:configurability}
\begin{lstlisting}[language=Python, caption=Configuring State-Spaces with SocialGym 2.0., label=lst:configurability]
observations = [
  AgentGoalDist(),
  # OtherAgentGoalDist(),
  ...
]
\end{lstlisting}


    


\noindent Each agent has a discrete action space $\mathcal{U}^i$\footnote{Future versions will include extending to continuous actions spaces}, and observation function $\mathcal{O}^i$ that takes in $x^i_t \in \mathcal{X}_t$ to output a local observation vector $o^i_t \in \Omega^i$ where $ o^i_t = [{x^i_t}, \tilde x^o_t]^\top$, and a reward function $\mathcal{R}^i: (\mathcal{X}_t,\mathcal{U}^i)\longrightarrow\mathbb{R}$. The transition function is given by $\mathcal{T}:\mathcal{X}\times U \longrightarrow \mathcal{X}$, where $U := \mathcal{U}_1 \times \mathcal{U}_2 \times \ldots \times \mathcal{U}_k$. Each agent has a policy distribution $\pi^i: \Omega^i \longrightarrow \Delta(\mathcal{U}^i)$ that takes in the local observation $o^i_t \in \Omega^i$ and stochastically performs action $u^i_t \in \mathcal{U}^i $ to produce a trajectory $\Gamma^i = \Big( x^i_I, x^i_2, \ldots, x^i_G  \Big)$, where $x^i_{t+1} = g(x^i_t,o^i_t, u^i_t | u^i_t \sim \pi^i(o^i_t))$ and $g(\cdot)$ is a local planner. The environment is geometrically constrained if there exists at least one point in the navigation graph, $p_\textrm{common} = (p_x, p_y)$, such that $p_\textrm{common} \in \Gamma^i \ \forall \ i = 1,2,\ldots,k$. 


The MARL objective is to find the optimal joint policy $\Pi^* = (\pi^*_1, \pi^*_2, \ldots, \pi^*_k)$ such that,
\begin{equation}
    \Pi^* = \arg\max_{ (\pi_1, \pi_2, \ldots, \pi_k)} \sum_{i=1}^k \mathbb{E}_{\pi^i} \Big[ \sum_{t\geq 0} \gamma^t\mathcal{R}^i(x^i_t, u^i_t) | u^i_t \sim \pi^i(o^i_t) \Big]
    \label{eq: global_cost}
\end{equation}  

\subsection{Hyperparameters}

Unless otherwise stated, we use Stable-Baselines3 PPO with a step size of $4096$ and the MLP architecture for training (the rest of the policies are hyperparameters of the default).  We train for a total of $1.25$ million steps, where the first $35$ episodes have $3$ agents, the next $35$ have $4$, and the remaining episodes have $5$ agents.  After training, we evaluate each policy on $25$ trials for $3, 4, 5, 7$, and $10$ agent settings. Agents may observe other agents' positions ($x$ and $y$ local to their coordinate frame), other agents' velocities, their own distance to the goal, if they are in a collision, and if they have succeeded.  Our reward function penalizes each agent for every step they are not at the goal ($-1$), a reward is given when the goal is reached ($100$), a penalty when an agent collides ($-10$), and a variable reward for making progress towards the goal (delta from the previous location to the current).  If no agents have moved over a significant delta (total magnitude of $0.5$ meters in $100$ steps) the episode ends and all agents not at their goal are given a penalty ($-100,000$).  Finally, we use wrappers to sample new trajectories through the map at the end of each episode and use Stable Baselines3 \texttt{VecNormalize} wrapper to normalize the observation and reward space (we found this to be very important to achieve stable results). An example of a configuration file for one of our experiments can be found in Listing \ref{lst:example_config}.


  

\begin{table}[t]
    \centering
    \resizebox{\columnwidth}{!}{
    \begin{tabular}{crcccc}
         \toprule
         Scenario &Baseline& Avg. Length &  Coll. Rate &Stop Time& Max $\Delta$V  \\
         \midrule
         \multirow{5}{*}{OPEN}&CADRL~\cite{cadrl}&526&0.24&287&71 \\
                                    &CADRL(L)~\cite{cadrl-lstm}&585&0.52&325&117 \\
                                    &Enforced Order &407&0.32&253&46 \\
                                    &\cellcolor{green}Any Order &\cellcolor{green!25}382&\cellcolor{green!25}0.12&234&\cellcolor{green!25}46 \\
                                    &Only Local&435&0.88&\cellcolor{green!25}187&72 \\
                                    \cmidrule{2-6}
         \multirow{5}{*}{DOORWAY}&\cellcolor{green}CADRL~\cite{cadrl}&\cellcolor{green!25}347&\cellcolor{green!25}0.00&222&\cellcolor{green!25}13 \\
                                    &CADRL(L)~\cite{cadrl-lstm}&602&0.12&436&36 \\
                                    &Enforced Order &960&0.16&617&117 \\
                                    &Any Order &667&0.32&324&51 \\
                                    &Only Local&411&2.32&\cellcolor{green!25}166&30 \\
                                    \cmidrule{2-6}
        \multirow{5}{*}{HALLWAY}&CADRL~\cite{cadrl}&564&\cellcolor{green!25}0.04&340&47 \\
                                    &CADRL(L)~\cite{cadrl-lstm}&875&0.84&441&96 \\
                                    &Enforced Order &917&0.68&538&43 \\
                                    &Any Order &621&1.00&\cellcolor{green!25}197&69 \\
                                    &\cellcolor{green}Only Local&\cellcolor{green!25}482&2.80&283&\cellcolor{green!25}24 \\
                                    \cmidrule{2-6}
       \multirow{5}{*}{INTERSECTION}&CADRL~\cite{cadrl}&691&1.20&389&\cellcolor{green!25}28 \\
                                    &\cellcolor{green}CADRL(L)~\cite{cadrl-lstm}&\cellcolor{green!25}678&\cellcolor{green!25}0.32&267&127 \\
                                    &Enforced Order &697&0.80&\cellcolor{green!25}233&104 \\
                                    &Any Order &999&0.68&592&47 \\
                                    &Only Local&1245&2.48&640&97 \\
                                    \cmidrule{2-6}                                        
      \multirow{5}{*}{ROUNDABOUT}&CADRL~\cite{cadrl}&733&0.64&395&99 \\
                                    &CADRL(L)~\cite{cadrl-lstm}&730&0.24&400&56 \\
                                    &\cellcolor{green}Enforced Order &\cellcolor{green!25}352&\cellcolor{green!25}0.00&\cellcolor{green!25}21&273 \\
                                    &Any Order &804&0.32&462&\cellcolor{green!25}27 \\
                                    &Only Local&2112&2.16&1194&163 \\
                           
         \bottomrule
    \end{tabular}
    }
    \caption{\textbf{Benchmarking various MARL baselines:} We compare six baselines. CADRL and its LSTM variants are state-of-the-art RL-based navigation algorithms, Enforced Order and Any Order are sub-goal reward policies that encourage queue formation, and Only Local is an ablation method where only the local motion planner is used. Green indicates best. Dark green indicates the overall best performing baseline for that scenario. \textbf{Conclusion:} There is no clear ``optimal social navigation'' algorithm in social mini-games. \smtr can be used to benchmark a range of policies to find the best one for a specific mini-game.}
    \label{tab: main_exps}
    \vspace{-10pt}
\end{table}

\subsection{Benchmarking Social Navigation Algorithms}
We benchmark $5$ baseline social navigation policies for each of the social mini-games described in Section~\ref{sec: minigames}. They are CADRL, CADRL(L), Enforced Order, Any Order, and Only Local. CADRL~\cite{cadrl} and its LSTM-based variant, which we denote as CADRL(L), are state-of-the-art multi-agent social navigation methods. CADRL and CADRL(L) use a reward function where an agent is rewarded upon reaching the goal and penalized for getting too close or colliding with other agents as well as taking too long to reach the goal. We train these baselines using PettingZoo and Stable Baselines3~\cite{sb3} and report results across a range of social navigation metrics in Table~\ref{tab: main_exps}. 

Enforced Order and Any Order are baselines that encourage agents to engage in social behaviors such as queue formations. Finally, Only Local is an ablated baseline where we remove the high-level policy from the MARL interface, reducing it to a purely local multi-agent navigation baseline. We compare these baselines across several social navigation metrics, following the standard literature \cite{tsoi2022sean,biswas2022socnavbench,socialnav_survey,stone_nav_survey}, and present results in Table \ref{tab: main_exps}. The experimental observations suggest that there is no straightforward ``optimal social navigation'' algorithm in social mini-games. \smtr can be used to benchmark a range of policies to find the best one for a specific mini-game.
\begin{lstlisting}[language=Python, caption=Example configuration that was used in our experiments, label=lst:example_config]
  "num_agents": [[0, 3], [35, 4], [70, 5]],
  "eval_num_agents": [3, 4, 5, 7, 10],
  "train_length": 250000, # 250k x 5agents = 1.25m
  "ending_eval_trials": 25,
  "eval_frequency": 0,
  "intermediate_eval_trials": -1,
  "policy_algo_sb3_contrib": false,
  "policy_algo_name": "PPO",
  "policy_name": "MlpPolicy",
  "policy_algo_kwargs": {"n_steps":  4096},
  "monitor": false,

  "experiment_names": ["envs_door"],

  "run_name": "door/ao",
  "run_type": "AO",
  "device": "cuda:0",

  "other_velocities_obs": true,
  "agent_velocity_obs": true,

  "agent_velocity_ignore_theta": false,
  "other_velocities_ignore_theta": false,
  "other_poses_ignore_theta": false,
  "agent_pose_ignore_theta": false,

  "entropy_constant_penalty": -100000,
  "entropy_constant_penalty_only_not_finish": true,
\end{lstlisting}




\paragraph{Advantage of sub-goal rewards in constrained social navigation}

The social mini-games in the benchmark are subject to geometric constraints that can cause conflicts between robots' paths. To evaluate the compatibility of standard social navigation metrics with human social behavior, we introduced a reward function based on the concept of queue formation. This reward function rewards robots for following a specified order to enter and exit conflict zones. Two new baselines were introduced to evaluate the efficacy of the reward function: "Any Order" and "Enforced Order". Any Order assigns a reward if a robot successfully passes through a conflict zone, regardless of the order, while Enforced Order assigns a random but specific order for robots to enter and exit. The results of the baselines are presented in Table~\ref{tab: main_exps} and Table~\ref{tab: general}.

The results show a mismatch between the standard social navigation metrics and success rates for different environments. In the Open and Roundabout scenarios, the sub-goals (Any Order and Enforced Order) have the greatest impact (Table~\ref{tab: main_exps}), but their success rates are low (Table~\ref{tab: general}).  While in the Intersection, Doorway, and Hallway environments, the success rates of the sub-goals are high but the metrics are poor. Although the policies may have learned a very social trait that is useful in environments requiring line formation, the standard metrics would not reflect this skill.  This suggests that the current metrics used to evaluate social navigation are insufficient and need to be improved. Thus, a goal for future work is to provide better extensibility of the evaluation metrics and to enhance the sub-goal models, Any Order and Enforced Order, for better results.

\begin{table}[t]
    \centering
    \resizebox{\columnwidth}{!}{
    \begin{tabular}{crabdecc}
         \toprule
         Scenario &Baseline& $\bm{3}$\textbf{A} &  $\bm{4}$\textbf{A} & $\bm{5}$\textbf{A} & $\bm{7}$\textbf{A} & $\bm{10}$\textbf{A} & \textbf{Avg}. \\
         \midrule
         \multirow{6}{*}{OPEN}&CADRL~\cite{cadrl}&32&28&8&0&0&7 \\
        &CADRL-LSTM~\cite{cadrl-lstm}&60&44&52&36&4&39 \\
    &Enforced Order &24&12&12&0&0&10 \\
                        &Any Order &16&12&8&8&0&9 \\
                        &Only Local&56&32&36&16&0&28 \\
                        \cmidrule{2-8}
\multirow{6}{*}{DOORWAY}&CADRL~\cite{cadrl}&28&32&16&8&0&17 \\
                        &CADRL-LSTM~\cite{cadrl-lstm}&8&0&4&12&16&8\\
                        &Enforced Order &64&44&28&8&4&30\\
                        &Any Order &72&84&64&24&4&50 \\
                        &Only Local&48&4&0&0&0&10 \\
                        \cmidrule{2-8}
\multirow{6}{*}{HALLWAY}&CADRL~\cite{cadrl}&24&24&12&4&0&13 \\
                        &CADRL-LSTM~\cite{cadrl-lstm}&68&40&12&0&0&24 \\
                        &Enforced Order &40&48&28&16&0&26 \\
                        &Any Order &44&28&16&0&0&18 \\
                        &Only Local&4&0&0&0&0&1 \\
                        \cmidrule{2-8}
\multirow{6}{*}{INTERSECTION}&CADRL~\cite{cadrl}&60&20&20&4&4&22 \\
                        &CADRL-LSTM~\cite{cadrl-lstm}&24&28&12&20&8&18 \\
                        &Enforced Order &80&56&44&28&0&42 \\
                        &Any Order &52&28&24&4&4&22 \\
                        &Only Local&32&16&0&0&0&10 \\
                        \cmidrule{2-8}                             
\multirow{6}{*}{ROUNDABOUT}&CADRL~\cite{cadrl}&24&20&8&0&0&10 \\
                        &CADRL-LSTM~\cite{cadrl-lstm}&40&8&0&8&0&11 \\
                        &Enforced Order &0&0&0&0&0&0 \\
                        &Any Order &16&20&12&0&4&10 \\
                        &Only Local&4&0&0&0&0&1 \\
                               
\bottomrule
\end{tabular}
}
\caption{\textbf{Success rates when generalizing from easier to harder environments:} We train policies with up to $4$ agents and test in environments with number of agents ranging from $2$ to $10$. Darker shades indicate better performance. \textbf{Conclusion:} Social navigation is harder as density of agents increases.}
\label{tab: general}
\end{table}

\begin{table*}[t]
    \centering
    \resizebox{\textwidth}{!}{
    \begin{tabular}{ccccccccc}
 \toprule
   & (MLP) Arch.&(\cmark) Goal Dist. & (\cmark) Collision Penalty& (\cmark) Existence Penalty & (\cmark) Entropy Penalty&(\xmark) Entropy Reward Multiplier & (\cmark) Velocities &(\cmark) Thetas \\
 \midrule
  Config&LSTM&Clipped&\xmark&\xmark&\xmark&\cmark&\xmark&\xmark \\
 Success Rate&76&0&84&44&52&20&0&0 \\
 \bottomrule
    \end{tabular}
    }
    \caption{\textbf{Success rates when testing different observation and reward configurations:} (\xmark) or (\cmark) in the header row indicate whether the corresponding reward, penalty, or observation was excluded or included in the baseline PPO policy. The ``config'' row then indicates how the configuration changed, and final row indicates the percentage of seed runs that performed better than the baseline.}
    \label{tab: obs_rew_config}
    \vspace{-10pt}
\end{table*}
\paragraph{More Complex Policies}

In our benchmarks, we evaluated the effectiveness of the Long Short-Term Memory with Proximal Policy Optimization (LSTM-PPO) from the SB3-Contrib library. The objective was to determine if collecting previous timesteps to form intermediate representations of state could improve performance in challenging scenarios. Although LSTM-PPO did not consistently outperform Proximal Policy Optimization (PPO) alone, it did demonstrate improved generalization to larger numbers of agents, sometimes achieving success in settings with up to ten agents. Our analysis of the performance of LSTM-PPO versus PPO, detailed in Table~\ref{tab: ppo}, specifically investigated the impact of the size of observations on policy updates. Our results suggest that policies that incorporate more complex representations of the environment, particularly those that encode temporal information, may be more effective. Consequently, as future work, we plan to incorporate state-of-the-art Multi-Agent Reinforcement Learning policies not currently supported by the native Stable Baselines library.


\paragraph{Role of the local planner in MARL-based navigation}

We also include another ablation study in which only the low-level planner and Ackermann steering are used. In this baseline, the multi-agent interface sends a "GO" command at all time steps, resulting in reactionary collision avoidance for each agent. Although this baseline can perform well in open environments with few agents, its limitations become evident as the number of agents increases or the environment becomes more complex. The results, as shown in Table~\ref{tab: general}, indicates that the "Only Local" baseline cannot successfully navigate environments with five or more agents, except for the open environment. Moreover, Table~\ref{tab: main_exps} reveals that "Only Local" has the highest average collision rate in an episode across all environments. This baseline demonstrates the importance of high-level planning in solving Social Navigation challenges in these mini-game environments and highlights the effectiveness of the discrete action space of "GO" and "STOP", despite its simplicity.


\paragraph{Experiments with different observation and reward functions}
\begin{figure}[t]
    \centering
\begin{subfigure}[h]{0.325\columnwidth}
    \includegraphics[width=\textwidth]{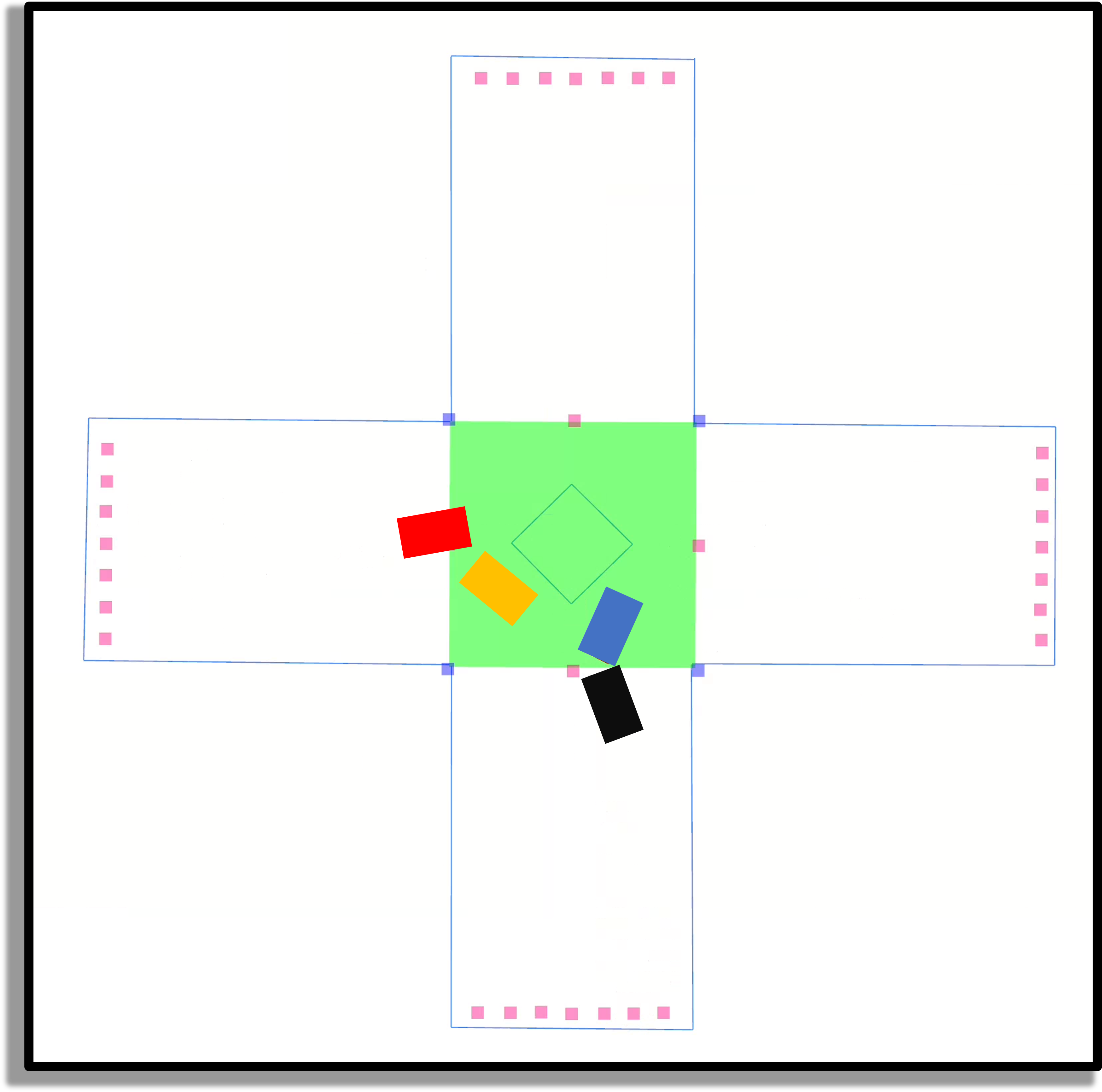}
    \caption{Entering the roundabout}
    \label{fig: inter1}
  \end{subfigure}
  \begin{subfigure}[h]{0.325\columnwidth}
    \includegraphics[width=\textwidth]{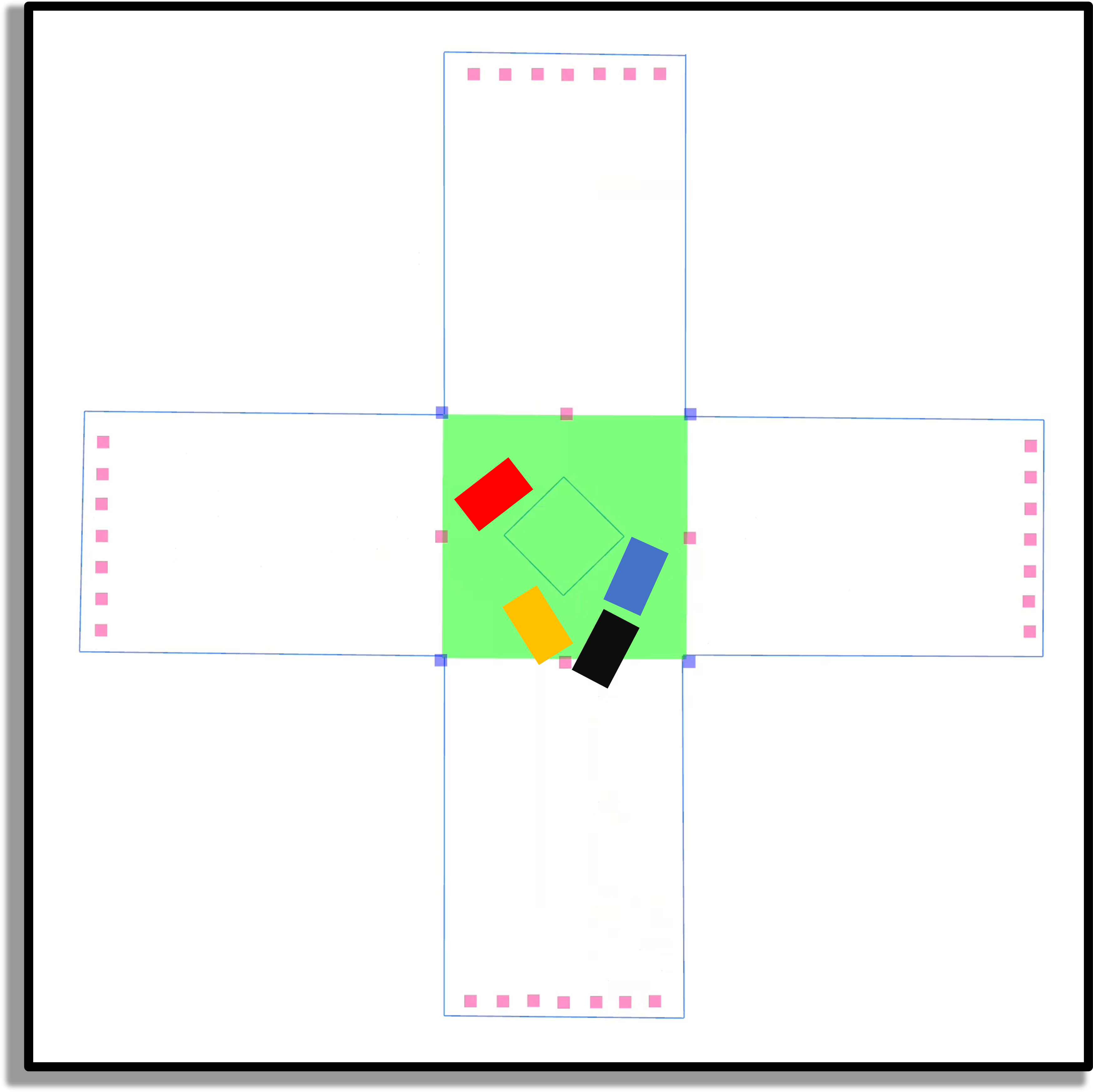}
    \caption{Navigating inside the roundabout}
    \label{fig: inter2}
  \end{subfigure}
  \begin{subfigure}[h]{0.325\columnwidth}
    \includegraphics[width=\textwidth]{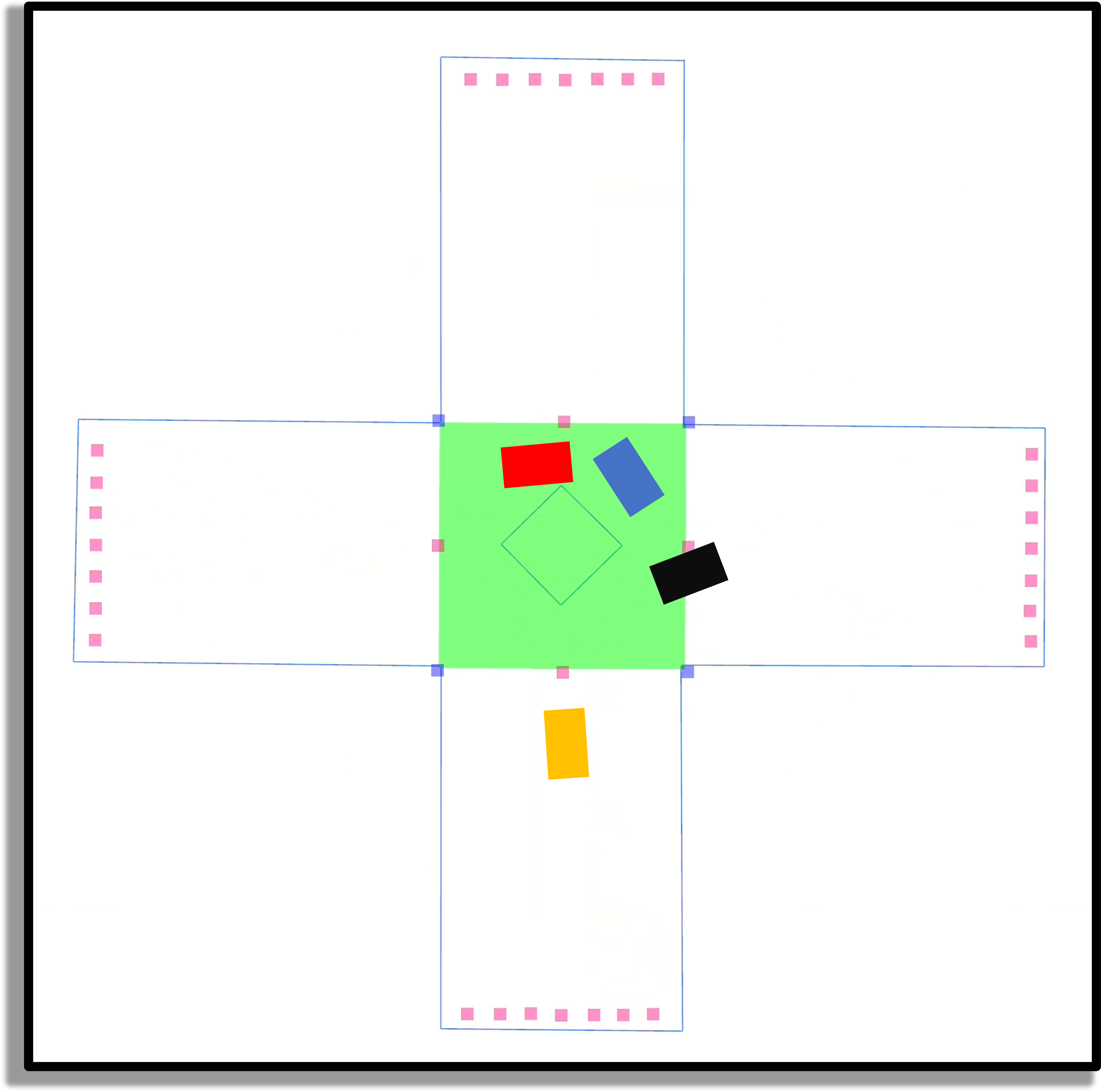}
    \caption{Exiting the roundabout}
    \label{fig: inter3}
  \end{subfigure}
  \caption{\textbf{Example of $4$ robots at an roundabout:} We trained a PPO policy in the roundabout scenario for $4$ agents. Agents navigate around each other in close quarters without colliding or slowing down. Blue lines represent physical walls and small pink squares represent starting positions.}
  \label{fig: intersection}
  \vspace{-10pt}
\end{figure}

Our final benchmark highlights the configurability of observation and reward functions in \smtr as a major advantage, enabling users to quickly and easily run multiple experiments to determine the optimal set of parameters. In Table~\ref{tab: obs_rew_config}, we present results for various configurations that we tested. The table indicates inclusion (\cmark) or exclusion (\xmark) of each reward or observation in each configuration. The baseline configuration is shown in the header row. These results are from the Intersection scenario with 4 agents.

The results demonstrate the significance of rapid testing and the ease with which different configurations can be explored in \smtr. The variability in success rates as a result of changes in the observation space and reward functions is notable. Surprisingly, removing the collision penalty appears to perform best in these experiments, although further investigation is required to fully understand this discrepancy. Our evidence suggests that agents with the collision penalty scale better to larger numbers of agents. Although we did not exhaustively test all possible configurations, these results demonstrate the importance of being able to quickly and easily explore different combinations of parameters.


\begin{table}[t]
    \centering
    \resizebox{.9\columnwidth}{!}{
    \begin{tabular}{ccccccc}
 \toprule
 & Arch. & \multicolumn{5}{c}{Batch Size}\\
 \midrule
 Config &MLP&\cellcolor{green!25}$128$&$256$&$512$&$1024$&$2048$\\
Success Rate&&52&32&8&36&24 \\
   \cmidrule{2-7}
 Config &LSTM&$128$&\cellcolor{green!25}$256$&$512$&$1024$&$2048$ \\
 Success Rate&&56&76&4&0&32 \\
 \bottomrule
    \end{tabular}
    }
    \caption{\textbf{Success rates of hyper-parameter Tuning:} The baseline consists of Goal Dist., Collision Penalty, Existence Penalty, Entropy Penalty, Entropy Reward Multiplier, observed velocities and headings. Green indicates best performance.}
    \label{tab: ppo}
    \vspace{-10pt}
\end{table}





\section{Conclusion, Limitations and Future Development Plans}
\label{sec: future_work}

In conclusion, this paper presents \smtr, a multi-agent navigation simulator designed to address the need for a realistic and challenging environment for social robot navigation research. The simulator provides a comprehensive solution to support research in this field, including a user-friendly interface, pre-packaged docker container and wrapper around PettingZoo's MARL library, as well as customizable observation and reward functions. Finally, we hope that the benchmarking of various social navigation algorithms demonstrates the potential of \smtr to advance the state of the art in this field.

However, \smtr has certain limitations that are currently being addressed, including constraints on CPU resources and lack of optimization for multi-threading and parallel processing. Additionally, parallel environments are not yet supported. Beyond addressing these limitations, we also plan to enhance the practicality of our platform by introducing continuous actions and state-of-the-art MARL algorithms in the multi-agent Gym interface, bringing it in line with cutting-edge advancements such as the work described in \cite{long2018towards}. Furthermore, we aim to provide more flexibility and control through variable observation vectors and streamline the configuration process through a unified file.


\IEEEpeerreviewmaketitle

{\footnotesize
\bibliographystyle{ieeetr}
\bibliography{refs}}

\begin{thebibliography}{10}

\bibitem{warehouse}
P.~R. Wurman, R.~D'Andrea, and M.~Mountz, ``Coordinating hundreds of
  cooperative, autonomous vehicles in warehouses,'' {\em AI magazine}, vol.~29,
  no.~1, pp.~9--9, 2008.

\bibitem{airport}
R.~Morris, C.~S. Pasareanu, K.~Luckow, W.~Malik, H.~Ma, T.~S. Kumar, and
  S.~Koenig, ``Planning, scheduling and monitoring for airport surface
  operations,'' in {\em Workshops at the Thirtieth AAAI Conference on
  Artificial Intelligence}, 2016.

\bibitem{pettingzoo}
J.~Terry, B.~Black, N.~Grammel, M.~Jayakumar, A.~Hari, R.~Sullivan, L.~S.
  Santos, C.~Dieffendahl, C.~Horsch, R.~Perez-Vicente, {\em et~al.},
  ``Pettingzoo: Gym for multi-agent reinforcement learning,'' {\em Advances in
  Neural Information Processing Systems}, vol.~34, pp.~15032--15043, 2021.

\bibitem{social_forces}
D.~Helbing and P.~Molnar, ``Social force model for pedestrian dynamics,'' {\em
  Physical review E}, vol.~51, no.~5, p.~4282, 1995.

\bibitem{tsoi2022sean}
N.~Tsoi, A.~Xiang, P.~Yu, S.~S. Sohn, G.~Schwartz, S.~Ramesh, M.~Hussein, A.~W.
  Gupta, M.~Kapadia, and M.~V{\'a}zquez, ``Sean 2.0: Formalizing and generating
  social situations for robot navigation,'' {\em IEEE Robotics and Automation
  Letters}, vol.~7, no.~4, pp.~11047--11054, 2022.

\bibitem{crowdbot}
F.~Grzeskowiak, D.~Gonon, D.~Dugas, D.~Paez-Granados, J.~J. Chung, J.~Nieto,
  R.~Siegwart, A.~Billard, M.~Babel, and J.~Pettr{\'e}, ``Crowd against the
  machine: A simulation-based benchmark tool to evaluate and compare robot
  capabilities to navigate a human crowd,'' in {\em 2021 IEEE International
  Conference on Robotics and Automation (ICRA)}, pp.~3879--3885, IEEE, 2021.

\bibitem{juliani2018unity}
A.~Juliani, V.-P. Berges, E.~Teng, A.~Cohen, J.~Harper, C.~Elion, C.~Goy,
  Y.~Gao, H.~Henry, M.~Mattar, {\em et~al.}, ``Unity: A general platform for
  intelligent agents,'' {\em arXiv preprint arXiv:1809.02627}, 2018.

\bibitem{biswas2022socnavbench}
A.~Biswas, A.~Wang, G.~Silvera, A.~Steinfeld, and H.~Admoni, ``Socnavbench: A
  grounded simulation testing framework for evaluating social navigation,''
  {\em ACM Transactions on Human-Robot Interaction (THRI)}, vol.~11, no.~3,
  pp.~1--24, 2022.

\bibitem{crowdnav}
C.~Chen, Y.~Liu, S.~Kreiss, and A.~Alahi, ``Crowd-robot interaction:
  Crowd-aware robot navigation with attention-based deep reinforcement
  learning,'' in {\em 2019 International Conference on Robotics and Automation
  (ICRA)}, pp.~6015--6022, IEEE, 2019.

\bibitem{aroor2017mengeros}
A.~Aroor, S.~L. Esptein, and R.~Korpan, ``Mengeros: A crowd simulation tool for
  autonomous robot navigation,'' in {\em 2017 AAAI Fall Symposium Series},
  2017.

\bibitem{socialgym}
J.~Holtz and J.~Biswas, ``Socialgym: A framework for benchmarking social robot
  navigation,'' {\em arXiv preprint arXiv:2109.11011}, 2021.

\bibitem{orca}
J.~Van Den~Berg, S.~J. Guy, M.~Lin, and D.~Manocha, ``Reciprocal n-body
  collision avoidance,'' in {\em Robotics Research: The 14th International
  Symposium ISRR}, pp.~3--19, Springer, 2011.

\bibitem{fridovich2020efficient}
D.~Fridovich-Keil, E.~Ratner, L.~Peters, A.~D. Dragan, and C.~J. Tomlin,
  ``Efficient iterative linear-quadratic approximations for nonlinear
  multi-player general-sum differential games,'' in {\em 2020 IEEE
  international conference on robotics and automation (ICRA)}, pp.~1475--1481,
  IEEE, 2020.

\bibitem{le2022algames}
S.~Le~Cleac’h, M.~Schwager, and Z.~Manchester, ``Algames: a fast augmented
  lagrangian solver for constrained dynamic games,'' {\em Autonomous Robots},
  vol.~46, no.~1, pp.~201--215, 2022.

\bibitem{chen2006convention}
N.~Chen, ``Convention over configuration,'' {\em h ttp://softwareengineering.
  vazexqi. com/files/pattern. htm l}, 2006.

\bibitem{sb3}
A.~Raffin, A.~Hill, A.~Gleave, A.~Kanervisto, M.~Ernestus, and N.~Dormann,
  ``Stable-baselines3: Reliable reinforcement learning implementations,'' {\em
  Journal of Machine Learning Research}, vol.~22, no.~268, pp.~1--8, 2021.

\bibitem{alphago}
D.~Silver, A.~Huang, C.~J. Maddison, A.~Guez, L.~Sifre, G.~Van Den~Driessche,
  J.~Schrittwieser, I.~Antonoglou, V.~Panneershelvam, M.~Lanctot, {\em et~al.},
  ``Mastering the game of go with deep neural networks and tree search,'' {\em
  nature}, vol.~529, no.~7587, pp.~484--489, 2016.

\bibitem{chess}
D.~Silver, T.~Hubert, J.~Schrittwieser, I.~Antonoglou, M.~Lai, A.~Guez,
  M.~Lanctot, L.~Sifre, D.~Kumaran, T.~Graepel, {\em et~al.}, ``Mastering chess
  and shogi by self-play with a general reinforcement learning algorithm,''
  {\em arXiv preprint arXiv:1712.01815}, 2017.

\bibitem{poker}
N.~Brown and T.~Sandholm, ``Superhuman ai for multiplayer poker,'' {\em
  Science}, vol.~365, no.~6456, pp.~885--890, 2019.

\bibitem{berner2019dota}
C.~Berner, G.~Brockman, B.~Chan, V.~Cheung, P.~D{\k{e}}biak, C.~Dennison,
  D.~Farhi, Q.~Fischer, S.~Hashme, C.~Hesse, {\em et~al.}, ``Dota 2 with large
  scale deep reinforcement learning,'' {\em arXiv preprint arXiv:1912.06680},
  2019.

\bibitem{vinyals2017starcraft}
O.~Vinyals, T.~Ewalds, S.~Bartunov, P.~Georgiev, A.~S. Vezhnevets, M.~Yeo,
  A.~Makhzani, H.~K{\"u}ttler, J.~Agapiou, J.~Schrittwieser, {\em et~al.},
  ``Starcraft ii: A new challenge for reinforcement learning,'' {\em arXiv
  preprint arXiv:1708.04782}, 2017.

\bibitem{gtsophy}
P.~R. Wurman, S.~Barrett, K.~Kawamoto, J.~MacGlashan, K.~Subramanian, T.~J.
  Walsh, R.~Capobianco, A.~Devlic, F.~Eckert, F.~Fuchs, {\em et~al.},
  ``Outracing champion gran turismo drivers with deep reinforcement learning,''
  {\em Nature}, vol.~602, no.~7896, pp.~223--228, 2022.

\bibitem{gt-marl-survey}
Y.~Yang and J.~Wang, ``An overview of multi-agent reinforcement learning from
  game theoretical perspective,'' {\em arXiv preprint arXiv:2011.00583}, 2020.

\bibitem{weerakoon2022terp}
K.~Weerakoon, A.~J. Sathyamoorthy, U.~Patel, and D.~Manocha, ``Terp: Reliable
  planning in uneven outdoor environments using deep reinforcement learning,''
  in {\em 2022 International Conference on Robotics and Automation (ICRA)},
  pp.~9447--9453, IEEE, 2022.

\bibitem{sathyamoorthy2022terrapn}
A.~J. Sathyamoorthy, K.~Weerakoon, T.~Guan, J.~Liang, and D.~Manocha,
  ``Terrapn: Unstructured terrain navigation using online self-supervised
  learning,'' in {\em 2022 IEEE/RSJ International Conference on Intelligent
  Robots and Systems (IROS)}, pp.~7197--7204, IEEE, 2022.

\bibitem{weerakoon2022graspe}
K.~Weerakoon, A.~J. Sathyamoorthy, J.~Liang, T.~Guan, U.~Patel, and D.~Manocha,
  ``Graspe: Graph based multimodal fusion for robot navigation in unstructured
  outdoor environments,'' {\em arXiv preprint arXiv:2209.05722}, 2022.

\bibitem{guan2022ga}
T.~Guan, D.~Kothandaraman, R.~Chandra, A.~J. Sathyamoorthy, K.~Weerakoon, and
  D.~Manocha, ``Ga-nav: Efficient terrain segmentation for robot navigation in
  unstructured outdoor environments,'' {\em IEEE Robotics and Automation
  Letters}, vol.~7, no.~3, pp.~8138--8145, 2022.

\bibitem{karnan2022vi}
H.~Karnan, K.~S. Sikand, P.~Atreya, S.~Rabiee, X.~Xiao, G.~Warnell, P.~Stone,
  and J.~Biswas, ``Vi-ikd: High-speed accurate off-road navigation using
  learned visual-inertial inverse kinodynamics,'' in {\em 2022 IEEE/RSJ
  International Conference on Intelligent Robots and Systems (IROS)},
  pp.~3294--3301, IEEE, 2022.

\bibitem{sathyamoorthy2020frozone}
A.~J. Sathyamoorthy, U.~Patel, T.~Guan, and D.~Manocha, ``Frozone:
  Freezing-free, pedestrian-friendly navigation in human crowds,'' {\em IEEE
  Robotics and Automation Letters}, vol.~5, no.~3, pp.~4352--4359, 2020.

\bibitem{sathyamoorthy2020densecavoid}
A.~J. Sathyamoorthy, J.~Liang, U.~Patel, T.~Guan, R.~Chandra, and D.~Manocha,
  ``Densecavoid: Real-time navigation in dense crowds using anticipatory
  behaviors,'' in {\em 2020 IEEE International Conference on Robotics and
  Automation (ICRA)}, pp.~11345--11352, IEEE, 2020.

\bibitem{sathyamoorthy2021comet}
A.~J. Sathyamoorthy, U.~Patel, M.~Paul, N.~K.~S. Kumar, Y.~Savle, and
  D.~Manocha, ``Comet: Modeling group cohesion for socially compliant robot
  navigation in crowded scenes,'' {\em IEEE Robotics and Automation Letters},
  vol.~7, no.~2, pp.~1008--1015, 2021.

\bibitem{cadrl}
Y.~F. Chen, M.~Liu, M.~Everett, and J.~P. How, ``Decentralized
  non-communicating multiagent collision avoidance with deep reinforcement
  learning,'' in {\em 2017 IEEE international conference on robotics and
  automation (ICRA)}, pp.~285--292, IEEE, 2017.

\bibitem{cadrl-lstm}
M.~Everett, Y.~F. Chen, and J.~P. How, ``Collision avoidance in pedestrian-rich
  environments with deep reinforcement learning,'' {\em IEEE Access}, vol.~9,
  pp.~10357--10377, 2021.

\bibitem{park2012robot}
J.~J. Park, C.~Johnson, and B.~Kuipers, ``Robot navigation with model
  predictive equilibrium point control,'' in {\em 2012 IEEE/RSJ International
  Conference on Intelligent Robots and Systems}, pp.~4945--4952, IEEE, 2012.

\bibitem{atreya2022high}
P.~Atreya, H.~Karnan, K.~S. Sikand, X.~Xiao, S.~Rabiee, and J.~Biswas,
  ``High-speed accurate robot control using learned forward kinodynamics and
  non-linear least squares optimization,'' in {\em 2022 IEEE/RSJ International
  Conference on Intelligent Robots and Systems (IROS)}, pp.~11789--11795, IEEE,
  2022.

\bibitem{wei2022steady}
J.~Wei, J.~Holtz, I.~Dillig, and J.~Biswas, ``Steady: Simultaneous state
  estimation and dynamics learning from indirect observations,'' in {\em 2022
  IEEE/RSJ International Conference on Intelligent Robots and Systems (IROS)},
  pp.~6593--6599, IEEE, 2022.

\bibitem{chandra2022gameplan}
R.~Chandra and D.~Manocha, ``Gameplan: Game-theoretic multi-agent planning with
  human drivers at intersections, roundabouts, and merging,'' {\em IEEE
  Robotics and Automation Letters}, vol.~7, no.~2, pp.~2676--2683, 2022.

\bibitem{chandra2020cmetric}
R.~Chandra, U.~Bhattacharya, T.~Mittal, A.~Bera, and D.~Manocha, ``Cmetric: A
  driving behavior measure using centrality functions,'' in {\em 2020 IEEE/RSJ
  International Conference on Intelligent Robots and Systems (IROS)},
  pp.~2035--2042, IEEE, 2020.

\bibitem{chandra_thesis}
R.~Chandra, {\em Towards Autonomous Driving in Dense, Heterogeneous, and
  Unstructured Traffic}.
\newblock PhD thesis, University of Maryland, College Park, 2022.

\bibitem{gameopt}
N.~Suriyarachchi, R.~Chandra, J.~S. Baras, and D.~Manocha, ``Gameopt: Optimal
  real-time multi-agent planning and control for dynamic intersections,'' in
  {\em 2022 IEEE 25th International Conference on Intelligent Transportation
  Systems (ITSC)}, pp.~2599--2606, IEEE, 2022.

\bibitem{socialmapf}
R.~Chandra, R.~Maligi, A.~Anantula, and J.~Biswas, ``Socialmapf: Optimal and
  efficient multi-agent path finding with strategic agents for social
  navigation,'' {\em arXiv preprint arXiv:2210.08390}, 2022.

\bibitem{socialnav_survey}
C.~Mavrogiannis, F.~Baldini, A.~Wang, D.~Zhao, P.~Trautman, A.~Steinfeld, and
  J.~Oh, ``Core challenges of social robot navigation: A survey,'' {\em arXiv
  preprint arXiv:2103.05668}, 2021.

\bibitem{stone_nav_survey}
R.~Mirsky, X.~Xiao, J.~Hart, and P.~Stone, ``Prevention and resolution of
  conflicts in social navigation--a survey,'' {\em arXiv preprint
  arXiv:2106.12113}, 2021.

\bibitem{Vector_Display}
U.~AMRL, ``Ut vector display.''
  \url{https://github.com/ut-amrl/vector_display}, 2021.

\bibitem{posg}
E.~A. Hansen, D.~S. Bernstein, and S.~Zilberstein, ``Dynamic programming for
  partially observable stochastic games,'' in {\em AAAI}, vol.~4, pp.~709--715,
  2004.

\bibitem{long2018towards}
P.~Long, T.~Fan, X.~Liao, W.~Liu, H.~Zhang, and J.~Pan, ``Towards optimally
  decentralized multi-robot collision avoidance via deep reinforcement
  learning,'' in {\em 2018 IEEE international conference on robotics and
  automation (ICRA)}, pp.~6252--6259, IEEE, 2018.

\end{thebibliography}
\end{document}